\documentclass[twoside]{article}

\RequirePackage[loading]{tracefnt}

\usepackage[accepted]{aistats2022}
\usepackage{graphicx}
\usepackage[round]{natbib}
\renewcommand{\bibname}{References}

\usepackage{hyperref}       
\usepackage{url}            
\usepackage{xcolor}         

\usepackage{booktabs}   
\usepackage{amsmath}    
\usepackage{amssymb}    
\usepackage{amsthm}     
\usepackage{booktabs}   
\usepackage{etoolbox}   
\usepackage{graphicx}   
\usepackage{hyperref}   
\usepackage{microtype}  
\usepackage{mleftright} 
\usepackage{multirow}   
\usepackage{nicefrac}   
\usepackage{paralist}   
\usepackage{siunitx}    
\usepackage{subcaption} 
\usepackage{tikz}       
\usepackage{pgfplots}   
\usepackage{soul}       
\usepackage{tabu}       
\usepackage{wrapfig}    
\usepackage{xspace}     
\usepackage[symbol]{footmisc}

\graphicspath{{..}}

\begin{document}

%

%

\theoremstyle{definition}
\newtheorem{definition}{Definition}[section]

\newtheorem{theorem}{Theorem}[section]
\newtheorem{corollary}{Corollary}[theorem]
\newtheorem{lemma}[theorem]{Lemma}
\newtheorem{assumption}[theorem]{Assumption}

\newcommand{\method}{CF-ODE\xspace}

\theoremstyle{remark}
\newtheorem*{remark}{Remark}

\renewcommand{\thefootnote}{\fnsymbol{footnote}}

\addtocounter{footnote}{0}

\twocolumn[\aistatstitle{Predicting the impact of treatments over time with uncertainty aware neural differential equations.}

\aistatsauthor{ Edward De Brouwer\footnotemark \And Javier Gonz\'{a}lez Hern\'{a}ndez \And  Stephanie Hyland }

\aistatsaddress{ ESAT-STADIUS \\ KU Leuven \And  Microsoft Research \\ Cambridge, UK \And Microsoft Research \\ Cambridge, UK } ]

\footnotetext{Work done during an internship at Microsoft Research.}

\begin{abstract}
Predicting the impact of treatments from observational data only still represents a major challenge despite recent significant advances in time series modeling. Treatment assignments are usually correlated with the predictors of the response, resulting in a lack of data support for counterfactual predictions and therefore in poor quality estimates. Developments in causal inference have lead to methods addressing this confounding by requiring a minimum level of overlap. However, overlap is difficult to assess and usually not satisfied in practice. 
In this work, we propose Counterfactual ODE (\method), a novel method to predict the impact of treatments continuously over time using Neural Ordinary Differential Equations equipped with uncertainty estimates. This allows to specifically assess which treatment outcomes can be reliably predicted. We demonstrate over several longitudinal data sets that \method provides more accurate predictions and more reliable uncertainty estimates than previously available methods.
\end{abstract}

\section{INTRODUCTION}

\begin{figure}[ht]
    \centering
    \includegraphics[width=0.38\textwidth]{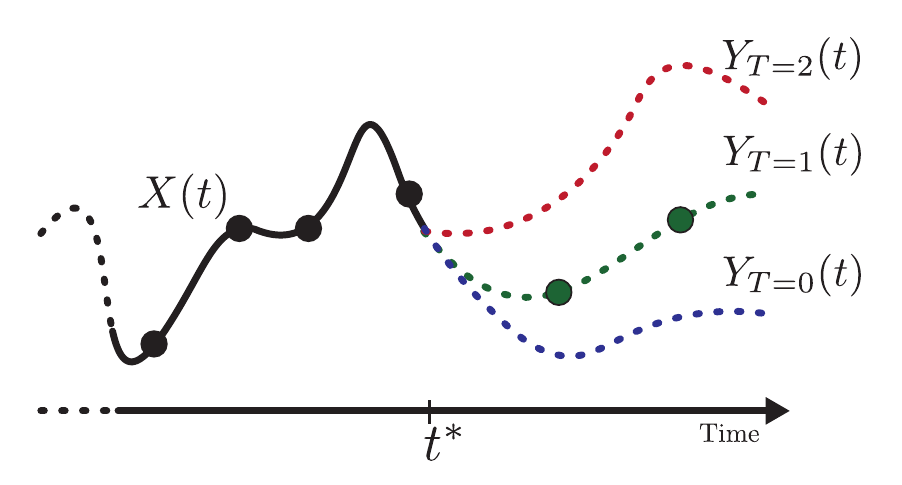}
  \caption{Based on available trajectory information $X(t)$, we aim at predicting in continuous time the potential outcomes of applying treatment regime $T$ at time $t^*$ (dotted lines). As with the fundamental problem of causal inference, a single outcome is available for each instance in the dataset (solid green dots).}
  \label{fig:illustatrive_example}
\end{figure}
   
The ability to forecast the impact of interventions on future outcomes is one of the main motivations of the scientific endeavour. To achieve this goal, the preferred strategy consists of experimentation by intervening on the process of interest. 
However, for practical or ethical reasons, many processes cannot be intervened on and this limitation has led to the successful development of tools for causal discovery and causal inference from observational data alone \citep{hernan2010causal,pearl2009causal}. One common example, and our main motivation in this work, consists of predicting the individual impact of medical interventions on patients. As clinical trials are long and expensive, the ability to predict individual treatment effects from observational data alone is particularly attractive. 

Based on observational clinical data, where treatments have been assigned to patients based on clinicians' judgement, we aim to predict the longitudinal progression of the disease course of individual patients depending on the intended treatment scenario. However, confounding, defined as the dependence of the treatment assignments on predictors of the future outcomes, can lead to biased estimates if not properly addressed \citep{bica2021real}. In this case, the cohorts of treated and non-treated patients are expected to be statistically different in the distribution of the predictors of the outcomes \citep{jesson2020identifying}. Crucially, this implies a distribution shift between the \emph{factual} (the observed treatment-outcome pairs) and the \emph{counterfactual} (when the treatment assignment is different than the one observed) distributions. Many methods in the literature have proposed strategies to address this distribution shift, usually requiring a significant level of positive overlap between treated and non-treated distributions, which is both difficult to test and seldom realized in practice \citep{oberst2020characterization,d2021overlap}. 

In contrast, rather than trying to improve the estimation of individual treatment effects (ITE) with a restrictive set of assumptions, we propose to learn a model to predict the different potential treatment outcomes over time with embedded uncertainties, reflecting the lack of data support in some regions of the predictors-treatment space \citep{jesson2020identifying}. We show that uncertainties in the prediction are crucial when it comes to the implementation of a treatment assignment recommendation system in (clinical) practice, where trust is of paramount importance. Crucially, it informs the decision maker as to which specific treatment outcomes can be reliably estimated. 

Another distinctive feature of temporal clinical time series is their irregular sampling times. Clinical data collection is usually driven by the underlying condition itself, leading to scattered data sampling times. This particularity, not unique to healthcare, led to the development of specific machine learning methods such as Neural Ordinary Differential Equations (ODE), capable to output predictions at arbitrary time points \citep{rubanova2019latent,de2019gru}.

Building upon those recent successes, we propose a latent neural ODE model equipped with uncertainty estimates to predict the individual impact of treatment assignments. We show that uncertainties in the ODE parameters can be encoded by reformulating the problem as a latent stochastic differential equation (SDE) model, relying on recently developed techniques \citep{li2020scalable,xu2021infinitely}. This formulation allows a flexible and efficient parametrization of the weights posterior probability distribution. Using datasets from cardiovascular system modeling and pharmacodynamics, we show that the uncertainties estimates reliably encode the error on the treatment effect estimator and provide a efficient way to detect patients for which an accurate estimation of treatment effect can be given.

\paragraph{Contributions} 
 
 \begin{itemize}
     \item We propose a novel uncertainty aware continuous time model that estimates the impact of interventions over time.
     \item To endow our model with uncertainties, we frame it as a latent neural stochastic differential equation that encodes the variational posterior of a latent neural ordinary differential equation model.
     \item We evaluate our approach on benchmarks from the literature on individualized treatment prediction in clinical machine learning.
 \end{itemize}
 
 \section{PROBLEM SETUP}

\subsection{Temporal Potential Outcomes Framework}

We aim at predicting longitudinal treatment outcomes based on historical observational data. We consider a set of $N$ multivariate time series $\mathcal{X} = \{ X_i(t) \in \mathbb{R}^{D}; i \in 1,...,N ; t \in \mathbf{t}^i \}$, sampled at arbitrary and potentially irregular time points $\mathbf{t}^i = \{t_0,...,t_{k_i} \}$ where $k_i$ stands for the number of observations of time series $X_i(t)$. We refer to $\mathcal{S}_{t'}(X_i) =  \{X_i(t) : t<t' \}$ as the set of available observations of $X_i(t)$ before $t'$. Some of those time series include a treatment assignment $T^*_i \in \{0,1\}$, where the star indicates the observed (\emph{i.e} factual) treatment assignment. In this case, we write the time of treatment assignment as $t_i^*$. 

We formulate our problem within the potential outcomes framework \citep{rubin1974estimating}. For each time series, we want to predict the potential outcomes $Y_{i,T}(t) : t\geq t_i^*$, the time series that would be observed when treatment $T$ is applied at time $t=t_i^*$. However, for each time series, only one potential outcome is observed, corresponding to the treatment that was actually given. In this case ($T=T^*_i$), $Y_{i,T}(t) = X_i(t) \quad \forall t>t_i^*$.
In our motivational patient trajectories example, the data would consist of $N$ patients for whom we observe $D$-dimensional time series. A treatment $T^*_i$ is then given at some time $t_i^*$ and the resulting treatment effect is observed over time ($Y_{i,T^*_i}(t)$). Based on this information, we want to be able to predict all potential outcomes on a new patient based on the available longitudinal data before treatment assignment ($\mathcal{S}_{t_i^*}(X_i)$). An example of time series (with $D=1$) and 3 potential treatment regimes is shown on Figure \ref{fig:illustatrive_example}. For sake of readability, we drop the $i$ subscript in the remaining of this text as the different time series are considered to be independent. 

\subsection{Dynamical System}
We consider the available temporal data $X(t)$ is modelled by a latent continuous time process $h(t)$ whose dynamics are characterized by an ordinary differential equation (ODE) as defined in Equation \ref{eq:ODE}. Conditioned on this latent process, observations X(t) are then obtained through an emission function $g(\cdot)$.

The impact of interventions on the dynamics are assumed to come from an external exogenous continuous temporal input $u_{T}(t)$ where $T$ indexes the treatment assignment. For a patient, this could represent the effect of an external intervention (\emph{e.g.} a drug) impacting their internal dynamics. To reflect the temporal causal structure of the exogenous intervention, we further require $u_{T}(t)=0 \quad \forall t<=0$. This leads to our first assumption about the data generating process. 

\begin{assumption}
All observations $X(t)$ and potential outcomes $Y(t)$ are driven by a common dynamical system characterized by an unknown ODE:
\label{ass:dynamical_system}
\end{assumption}

\begin{equation}
\begin{aligned}[b]
\frac{dh_T(t)}{dt} &= f(h_T(t),u_{T}(t-t^*)) \\
X(t) &\sim g(h_{T^*}(t))\\
Y_T(t) &\sim g(h_T(t)), \forall t\geq t^*
\end{aligned}
\label{eq:ODE}
\end{equation}

where $f(\cdot)$, $g(\cdot)$ and $u_T(\cdot)$ are unknown functions. Note that before the treatment assignment ($t<t^*$), all potential outcomes trivially coincide with the observed process $X(t)$.

\subsection{Treatment Assignment and Confounding}

\begin{figure}[hb]
    \centering
    \includegraphics[width=0.33\textwidth]{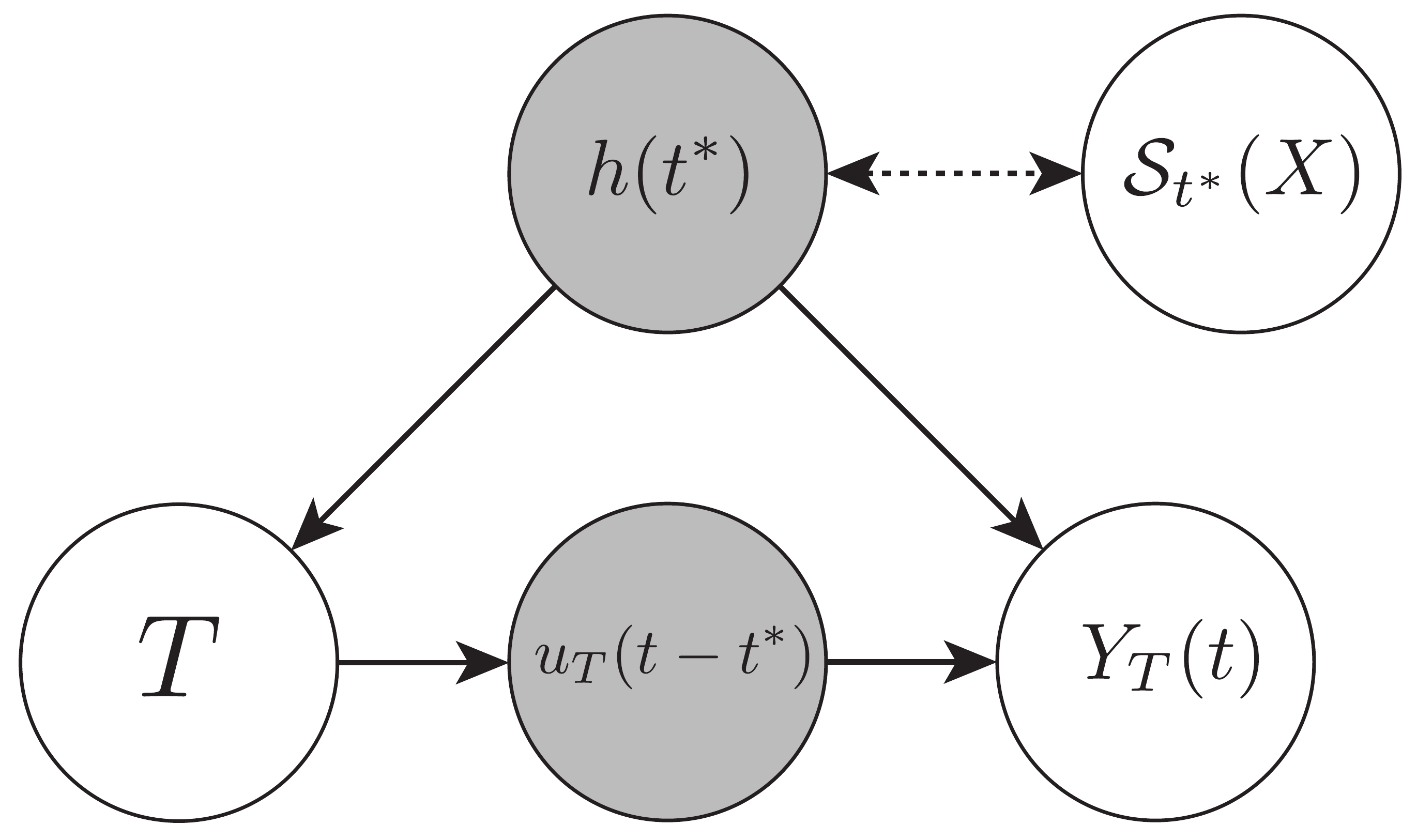}
  \caption{Graphical model representation of the problem setup. Greyed variables are unobserved. The link between $h(t^*)$ and $\mathcal{S}_{t^*}(X)$ embodies Assumption \ref{ass:confounder}. }
  \label{fig:graphical_model}
\end{figure}

The main challenge in causal inference from observational data arises because observed treatment assignments are not at random. They may depend on the past observations $X(t): t\leq t^*$ or on other variables that are not directly observed. This becomes problematic when the variables on which depends the treatment assignment are also predictive of the potential outcomes, a case usually referred to as \emph{confounding}. 

Without loss of generality, we consider that the treatment assignment at time $t=t^*$ depends on the latent process $h(t^*)$, giving the following propensity model:

\begin{assumption}
Conditioned on a treatment assignment time $t^*$, the probability of treatment assignment is generated as $T(t^*) \sim \tau(h(t^*))$.
\label{ass:treatment_assignment}
\end{assumption}

In this work, for sake of simplicity, the time of treatment $t^*$ is also supposed to be fixed and we consider binary treatments. Details on the more general case are given in Appendix \ref{app:no_treatment_label}.

As the treatment assignment depends on the \emph{unobserved} latent process $h(t)$, we require another assumption to ensure that we can control for all possible confounders:

\begin{assumption}
There exists a map $\phi (\cdot)$ between  the set of available measurements at the time of treatment ($\mathcal{S}_{t^*}(X)$) and the latent process at time of treatment ($h(t^*)$) such that, for all observed times series $X(t)$:\\
$\phi(\mathcal{S}_{t^*}(X)) = h(t^*)$.
\label{ass:confounder}
\end{assumption}

This assumption corresponds to the classical strong ignorability condition. 
Intuitively, it suggests that $\mathcal{S}_{t^*}(X)$ contains enough information to fully describe the state of the dynamical system at time $t=t^*$. As we will show in Section \ref{sec:model}, this assumption implies that by adjusting for $\mathcal{S}_{t^*}(X)$, we automatically adjust for all potential confounders. 
Importantly, because of the dynamic nature of time series, this assumption might in practice be less restrictive than the traditional strong ignorability assumption, as shown in more details in Appendix \ref{ass:confounder}. A graphical model summarizing the assumptions is presented on Figure \ref{fig:graphical_model}.

 \subsection{Overlap Of Treated Distributions}
 \label{sec:overlap}
 
 On top of the strong ignorability assumption, positive overlap between the distributions of the different treated groups is usually required in the potential outcomes framework. Unfortunately, this assumption is rarely satisfied, limiting the applicability of methods relying on it \citep{oberst2020characterization}. Therefore, in this work, we do not assume positive overlap between the various treated groups but rather model the lack of overlap by making our model uncertainty aware, as developed in Section \ref{sec:model}.
 

\section{MODEL}
\label{sec:model}

\subsection{Adjusting For Confounders}
To avoid biased estimates of the potential outcomes, our model should control for all 
confounders. Relying on previously defined assumptions, we can ensure that this is the case if we use $\mathcal{S}_{t^*}(X)$ as input in our model, as shown by the following lemma: 

\begin{lemma}
If assumptions \ref{ass:treatment_assignment} and \ref{ass:confounder} are met, then adjusting for $\mathcal{S}_{t^*}(X)$ controls for all possible confounders of $T$ and $Y$. 

\textit{Proof} \quad By the existence of a function such as in assumption \ref{ass:confounder}, conditioning on the $\mathcal{S}_{t^*}(X)$ is equivalent to conditioning on the $h(t^*)$. As all possible confounding between treatment assignment and outcomes go through $h(t^*)$, adjusting for $\mathcal{S}_{t^*}(X)$ controls for all possible confounders.
\label{lemma:confounder}
\end{lemma}

\subsection{Characterizing The Lack Of Overlap With Uncertainties}

As stated above, we do not explicitly assume sufficient overlap. Nevertheless, the absence of overlap can be characterized as it would result in more \emph{epistemic} uncertainty of the estimators in the region of poor coverage \citep{jesson2020identifying}.
In particular, for the same observed time series $X(t)$, the uncertainty about the predicted potential outcomes can vary based on the treatment assignment, depending of how often particular treatments are observed for similar time series in the dataset. The resulting uncertainty can then be used to inform where reliable treatment effect predictions can be made. Indeed, in the limit of a dataset where there is no overlap between treated on non-treated, nothing can be expected from counterfactual prediction as literally no datapoints are available to train the counterfactual model. Therefore, we want to equip our model with epistemic uncertainty estimates, to reflect the lack of data coverage in specific observation-treatment regions.

\subsection{Learning The Dynamics With Neural ODEs}

Building upon the formulation of the data generating process (Eq. \ref{eq:ODE}) and our result from Lemma \ref{lemma:confounder}, we propose to model the dynamics of the observed time series in two steps. We first encode the available measurements up until the treatment assignment time ($t^*$), aiming to recover the hidden process $h(t^*)$. We then integrate forward the hidden process $h(t)$ with the treatment-specific exogenous inputs ($u(t)$) using a Neural ODE to predict the potential outcomes $Y(t)$.

The encoder mapping the observed data points to the latent space is a neural network $\Phi$ with parameters $\phi$: 

\begin{align}
    h(t^*) = \Phi(\mathcal{S}_{t^*}(X))
    \label{eq:encoder}
\end{align}

The Neural ODE used to compute predictions of treatment outcomes $Y(t)$ follows the same structure as Equation \ref{eq:ODE} where we parametrize unknown functions 
with neural networks $f_{\theta_f}(\cdot)$, $g_{\theta_g}(\cdot)$ and $u_{T,\theta_u}(\cdot)$:

\begin{align}
\frac{dh_T(t)}{dt} &= f_{\theta_f}(h_T(t),u_{T,\theta_u}(t-t^*)) \nonumber\\
Y_T(t) &\sim g_{\theta_g}(h_T(t)), \forall t\geq t^*.
\label{eq:NODE}
\end{align}

In this work, we use multi-layer perceptrons for $f$,$g$ and $u$ with the size of the hidden layer equal to the size of the latent process and used a GRU network for $\Phi$.

\begin{figure*}[t]
\centering
   \includegraphics[width=1\linewidth]{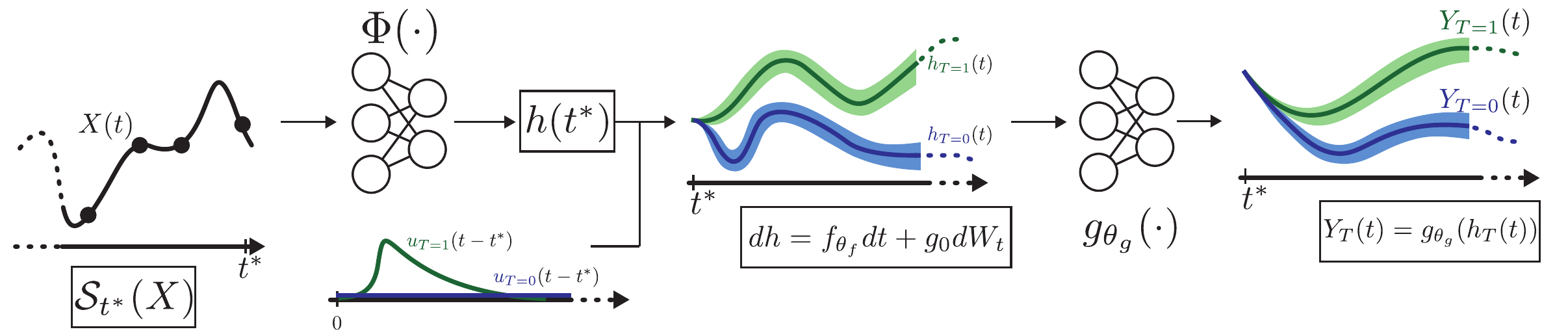}
  \caption{Overview of the \method architecture. Available temporal information ($\mathcal{S}_t(X)$) is mapped to the latent vector $h(t^*) = \Phi(\mathcal{S}_t^*(X))$. The latent process $h(t)$ is then integrated over time using a learnt Neural SDE. The impact of treatment is taken into account in the dynamics by learning an exogeneous treatment process $u_T(t)$. Finally, predictions of treatment effects in the observation space are obtained by applying a pointwise mapping $g_{\theta_g}(\cdot)$ on the hidden process.}
  \label{fig:architecture}
\end{figure*}

\subsection{Incorporating Uncertainties In ODEs Using Stochastic Differential Equations}

We propose to embed uncertainties in the parameters of the ODE. Adopting the formalism from Bayesian neural networks, we posit a prior on the weights of the neural ODE function 
: $\theta_f \sim P(\theta_f)$ and aim to estimate the posterior weight distribution given the available data $P(\theta_f \mid \mathcal{D})$ where $\mathcal{D} = \{ \mathcal{S}_{t^*},Y \}$. The weights being probabilistic, the ODE therefore effectively becomes a \emph{random} differential equation. The generating process of $h_T(t)$ then goes as

\begin{align*}
\theta_f &\sim P(\theta_f) \\
     h_T(t) &= h_T(t^*) + \int_{t^*}^{t}
    f(h_T(s),u_{T,\theta_u}(s-t^*), \theta) \cdot ds
\end{align*}

where we wrote the dependence on the weigths $\theta_f$ explicitly. 
This means that the prior on the weights $\theta_f$ entails a prior distribution on the process $h_T(t)\mid h_T(t^*)$. For brevity, we refer to this process as $\mathcal{H}$. Using the encoder-decoder decomposition of our model, we can derive a variational bound for the marginal probability of the available data $\mathcal{D}$, where we make the dependence on the treatment implicit:

\begin{align}
    \log(p(\mathcal{D})) &= \nonumber \\
    \log \int_{\mathcal{H}} \int_{\theta_f} &p(Y\mid \mathcal{H}) \cdot p(\mathcal{H} \mid \mathcal{S}_{t^*},\theta_f) \cdot p(\theta_f) \cdot d\theta_f \cdot d\mathcal{H}. \nonumber \\
    &\leq \mathbb{E}_{q_\theta(\mathcal{H}\mid \mathcal{S}_{t^*})} [ \log p(Y\mid \mathcal{H})] \nonumber \\
    &- KL_{q_\theta(\mathcal{H}\mid \mathcal{S}_{t^*})} (q_\theta(\mathcal{H}\mid \mathcal{S}_{t^*}) \mid \mid p_0(\mathcal{H}\mid \mathcal{S}_{t^*})) \nonumber \\
    \label{eq:variational}
\end{align}

where $p_0(\mathcal{H}\mid \mathcal{S}_{t^*})$ stands for the prior on the process $\mathcal{H}$.The process $\mathcal{H}$ being stochastic, we parametrize the variational approximation of the posterior distribution $q_\theta(\mathcal{H}\mid \mathcal{S}_{t^*})$ with a neural \emph{stochastic} differential equation (SDE) (in particular, a diffusion process),

\begin{align}
    q_\theta(\mathcal{H}\mid \mathcal{S}_{t^*}) &\sim \nonumber\\
    dh(t) &= f_{\theta_f}(h(t),u_{T,\theta_u}(t-t^*)) dt + g_{\phi_g}(h(t)) dW_t
    \label{eq:posterior_sde}
\end{align}

Where $dW_t$ stands for the multi-dimensional Wiener process. Similarly as in Neural ODEs, a Neural SDE is simply an SDE parametrized by neural networks.

The challenging part in optimizing the variational bound of Eq. \ref{eq:variational} resides in computing the KL divergence. Fortunately, this divergence becomes tractable if we choose the prior $p_0(\mathcal{H}\mid \mathcal{S}_{t^*})$ to be another diffusion process with same diffusion parameter $g_{\theta_g}$. As we don't learn the prior, this comes down to fixing the diffusion parameter, which might appear very restrictive. However, previous results from Neural SDE literature show that any posterior can be approximated arbitrarily closely by such a functional form given a sufficiently expressive drift process ($f_{\theta_f}$) \citep{boue1998variational,tzen2019neural,xu2021infinitely}. The prior on the process $\mathcal{H}$ then writes : 

\begin{align*}
    p_0(\mathcal{H}\mid \mathcal{S}_{t^*}) \sim dh(t) = f_0(h(t)) dt + g_0(h(t)) dW_t,
\end{align*}

The KL divergence is then tractable and is given by 

\begin{align*}
   KL_{q_\theta(\mathcal{H}\mid \mathcal{S}_{t^*})} &(q_\theta(\mathcal{H}\mid \mathcal{S}_{t^*}) \mid \mid p_0(\mathcal{H}\mid \mathcal{S}_{t^*}))  \\
   &= \mathbb{E}_{q_\theta(\mathcal{H}\mid \mathcal{S}_{t^*})} [\int_0^T \mid \mid u(t,\gamma) \mid \mid_2^2 dt] \\
   \text{where} \quad u(t,\gamma) &=  g_0(\mathcal{H}(t))^{-1} [f_{\theta_0}(\mathcal{H}(t))-f_0(\mathcal{H}(t))]
\end{align*}

We study the impact of the diffusion term $g_0$ (\emph{i.e.} with respect to a classical Neural ODE) in Appendix \ref{app:ablation}.

\subsection{\method}
\label{sec:the_method}

We are now ready to formalize the final form of our model. A graphical illustration is shown on Figure \ref{fig:architecture}. Using the encoder from Eq. \ref{eq:encoder}, we map the observed time series to the hidden space $h(t^*)$. We then reconstruct the potential outcomes by integrating the SDE from Equation \ref{eq:posterior_sde} and mapping from latent space to observation space using $\mu_{Y_T}(h_T(t)), \sigma_{Y_T}(h_T(t)) = g_{\theta_g}(h_T(t))$, where $\mu$ and $\sigma$ are the parameters of the predicted distribution of $Y_T(t)$ (assumed Gaussian in our experiments). The loss we optimize is then
\begin{align*}
    \mathcal{L}(\mathcal{D},\theta,\phi) &= \\ \mathbb{E}_{q_\theta(\mathcal{H}\mid \mathcal{S}_{t^*})} &[ \log p_{\theta}(Y\mid \mathcal{H})] \\
    -  KL_{q_\theta(\mathcal{H}\mid \mathcal{S}_{t^*})} &(q_\theta(\mathcal{H}\mid \mathcal{S}_{t^*}, \phi) \mid \mid p_0(\mathcal{H}\mid \mathcal{S}_{t^*}, \phi))
\end{align*}

which, remarkably, can be learned end to end using a differentiable SDE solver \citep{kidger2021neuralsde,li2020scalable}. In our experiments, we found that fixing $\sigma_X(h(t))$ as a hyperparameter helped the training and resulted in better performance. We provide an in-depth discussion of this effect and a corresponding ablation study in Appendix \ref{app:ablation}. More details about the variational bound derivations are given in Appendix \ref{app:long_proof}.

\section{EXPERIMENTS}

We evaluate our method on three datasets: a synthetic one, a dataset from cardiovascular modeling and one from pharmacodynamics. Full details about these datasets are given in Appendix \ref{app:datasets}. For each of them, we report the root mean-squared error (RMSE) obtained on a left-out test set for the factual distribution, the counterfactual (off-policy) distribution and the treatment effect (PEHE \citep{hill2011bayesian}). The Precision in Estimation of Heterogeneous Effect (PEHE) represents the ability to predict the size of the treatment effect (that is one treatment against another). The $PEHE$ between a treatment $T'$ and a treatment $T$ is given by $ \sqrt{((Y_{i,T}- Y_{i,T'})-(\hat{Y}_{i,T} - -\hat{Y}_{i,T'}))^2}$.

To assess the relevance of the uncertainty estimates, we also report the same metrics with only half of the test set samples as introduced in \citet{jesson2020identifying}. They describe different strategies for picking the retained samples. The first one is to keep samples with lowest predicted uncertainty. The second one is to keep sample according to the propensity score (unlikely predictors-treatment assignment pairs are discarded first). The last one is a random sample of the test set. The code for reproducing the experiments is available at \url{https://github.com/microsoft/cf-ode}.

\subsection{Baselines}

We compare our method against a set of state-of-the-art methods for the prediction of individual treatment effects. In particular, we compare against the counterfactual recurrent network (CRN) \citep{bica2020estimating}, Gaussian processes (GP) \citep{schulam2017reliable} and IMODE \citep{gwak2020neural}. As most of these methods do not model uncertainties and do not allow for irregular sampling (except for the GP), we use the (exact) propensity-based strategy and regularly sampled data in these cases. Our approach, in contrast supports irregular sampling. Further details about the implementations of the baselines can be found in Appendix \ref{app:baselines}. For sake of fair comparison, we used 
the same train-validation-test split for all methods. 

\subsection{Datasets}

\paragraph*{Harmonic Oscillator.}

We first demonstrate the performance of our approach on a synthetic dynamical system. We model the dynamics of a pendulum where the intervention consists of injecting energy into the system over time. 
Treatment assignments are coupled with the trajectories through a parameter $\gamma$ that controls the level of confounding.

 \begin{table*}[t]
  \caption{%
   Test RMSE for in-distribution, out-distribution and PEHE for the different methods. Rows with `50\%' refer to trimming the test set to those with either lowest uncertainty, or propensity scores closest to 0.5 (propensity). Best results and the ones that do not significantly differ from best are in bold.
  }
  \label{tab:results}
  \vskip 0.1in
  \centering
  \small
  \newcommand{\gr}{\rowfont{\color{gray}}}
  \newcommand{\NA}{---}
  \sisetup{
    detect-all           = true,
    table-format         = 2.1(2),
    separate-uncertainty = true,
    mode                 = math,
    table-text-alignment = center,
    tight-spacing,
  }
  \robustify\bfseries
  \renewrobustcmd{\bfseries}{\fontseries{b}\selectfont}
  \renewrobustcmd{\boldmath}{}
  \let\b\bfseries
  \setlength{\tabcolsep}{3.0pt}
  
  \begin{tabu}{llSSS}
    \toprule
    {Dataset} & \textsc{Method}     & {In-distribution RMSE} & {Out-distribution RMSE} & {PEHE}\\
    \midrule
    Oscillator & \method   &\b\num{0.04 \pm 0.01} & \b\num{0.06 \pm 0.01} & \b\num{0.07 \pm 0.01} \\
    & CFR &\num{0.11 \pm 0.01} & \num{0.24 \pm 0.02} & \num{0.28 \pm 0.02} \\
     & IMODE  &\num{0.30 \pm 0.04} & \num{0.34 \pm 0.04} & \num{0.49 \pm 0.08} \\
    & GP &\num{0.57 \pm 0.01} & \num{0.56 \pm 0.01} & \num{0.88 \pm 0.01} \\
     \midrule
     & \method  50\% (uncertainty) & \b\num{0.03 \pm 0.01} & \b\num{0.03 \pm 0.01} & \b\num{0.05 \pm 0.01} \\
    & \method  50\% (propensity)        &  \b\num{0.04 \pm 0.01} & \b\num{0.04 \pm 0.01} & \b\num{0.06 \pm 0.01} \\
    & CFR 50\% (propensity)   &\num{0.09 \pm 0.01} & \num{0.15 \pm 0.02} & \num{0.22 \pm 0.01} \\
    & IMODE 50\% (propensity)   &\num{0.30 \pm 0.04} & \num{0.34 \pm 0.04} & \num{0.47 \pm 0.08} \\
    & GP 50\% (uncertainty)&\num{0.57 \pm 0.01} & \num{0.57 \pm 0.01} & \num{0.86 \pm 0.03}\\
    \midrule
    \midrule
    Cardio-vascular & \method   &  \num{0.21 \pm 0.02} & \num{0.59 \pm 0.03} & \num{0.69 \pm 0.03} \\
      & CFR  &\num{0.26 \pm 0.02} & \num{0.90 \pm 0.04} & \num{0.98 \pm 0.03} \\
    & IMODE  & \b\num{0.16 \pm 0.02} & \b\num{0.38 \pm 0.02} & \b\num{0.46 \pm 0.02} \\
    & GP &\num{0.41 \pm 0.02} & \num{0.52 \pm 0.02} & \num{0.72 \pm 0.01} \\
    
     \midrule
     
     & \method  50\%  (uncertainty)       &  \b\num{0.11 \pm 0.02} & \b\num{0.39 \pm 0.03} & \num{0.69 \pm 0.06} \\
    & \method  50\%  (propensity)  &  \num{0.20 \pm 0.02} & \b\num{0.33 \pm 0.06} & \num{0.45 \pm 0.06} \\
    & CFR 50\% (propensity)   &\num{0.24 \pm 0.01} & \num{0.55 \pm 0.04} & \num{0.61 \pm 0.08} \\
    & IMODE 50\% (propensity)   &\num{0.16 \pm 0.02} & \b\num{0.36 \pm 0.02} & \b\num{0.35 \pm 0.02} \\
    & GP 50\% (uncertainty)   &\num{0.39 \pm 0.01} & \num{0.51 \pm 0.03} & \num{0.69 \pm 0.04} \\
    
    \midrule
    \midrule
    Dexamethasone & \method   &  \b\num{0.027 \pm 0.002} & \b\num{0.049 \pm 0.005} & \b\num{0.050 \pm 0.003} \\
      & CFR  &\num{0.038 \pm 0.003} & \num{0.066 \pm 0.001} & \num{0.084 \pm 0.001} \\
    & IMODE &\num{0.051 \pm 0.024} & \b\num{0.051 \pm 0.024} & \num{0.094 \pm 0.043} \\
    & GP &\num{0.635 \pm 0.090} & \num{0.619 \pm 0.086} & \num{0.086 \pm 0.012} \\
     \midrule
    & \method  50\% (uncertainty) &  \b\num{0.016 \pm 0.003} & \b\num{0.025 \pm 0.004} & \b\num{0.018 \pm 0.001} \\
    & \method  50\%  (propensity)       &  \num{0.026 \pm 0.005} & \num{0.038 \pm 0.005} & \num{0.044 \pm 0.006} \\
    & CFR 50\% (propensity)   &\num{0.026 \pm 0.002} & \num{0.057 \pm 0.006} & \num{0.088 \pm 0.006} \\
    & IMODE 50\% (propensity)   &\num{0.052 \pm 0.024} & \num{0.052 \pm 0.024} & \num{0.094 \pm 0.043} \\
    & GP 50\%  (uncertainty) &\num{0.620 \pm 0.062} & \num{0.607 \pm 0.059} & \num{0.083 \pm 0.011} \\
    
    \bottomrule
    
  \end{tabu}

\end{table*}

\paragraph*{Cardiovascular Model.}

We use a model of the cardiovascular system proposed in \citet{zenker2007inverse} and \citet{linial2021generative} to study the capacity of our model to learn the impact of fluids intake. Fluids are commonly administered for treating severe hypotension, but individual patients response is difficult to assess beforehand, making it a very pertinent case study.

\paragraph*{Pharmacodynamics Model.}

As a third example, we consider the pharmacodynamics of dexamethasone, a glucocorticoid drug used against COVID-19. The dynamical system is adapted from \citet{dai2021prototype} and \citet{qian2021integrating}. Based on time series from Type I IFNs and Cytotoxic T Cells measurements, we aim at predicting the level of Type I IFNs in response to dexamethasone.

\section{RESULTS}
\label{sec:results}

\begin{figure}[ht]
    \centering
    \includegraphics[width=0.42\textwidth]{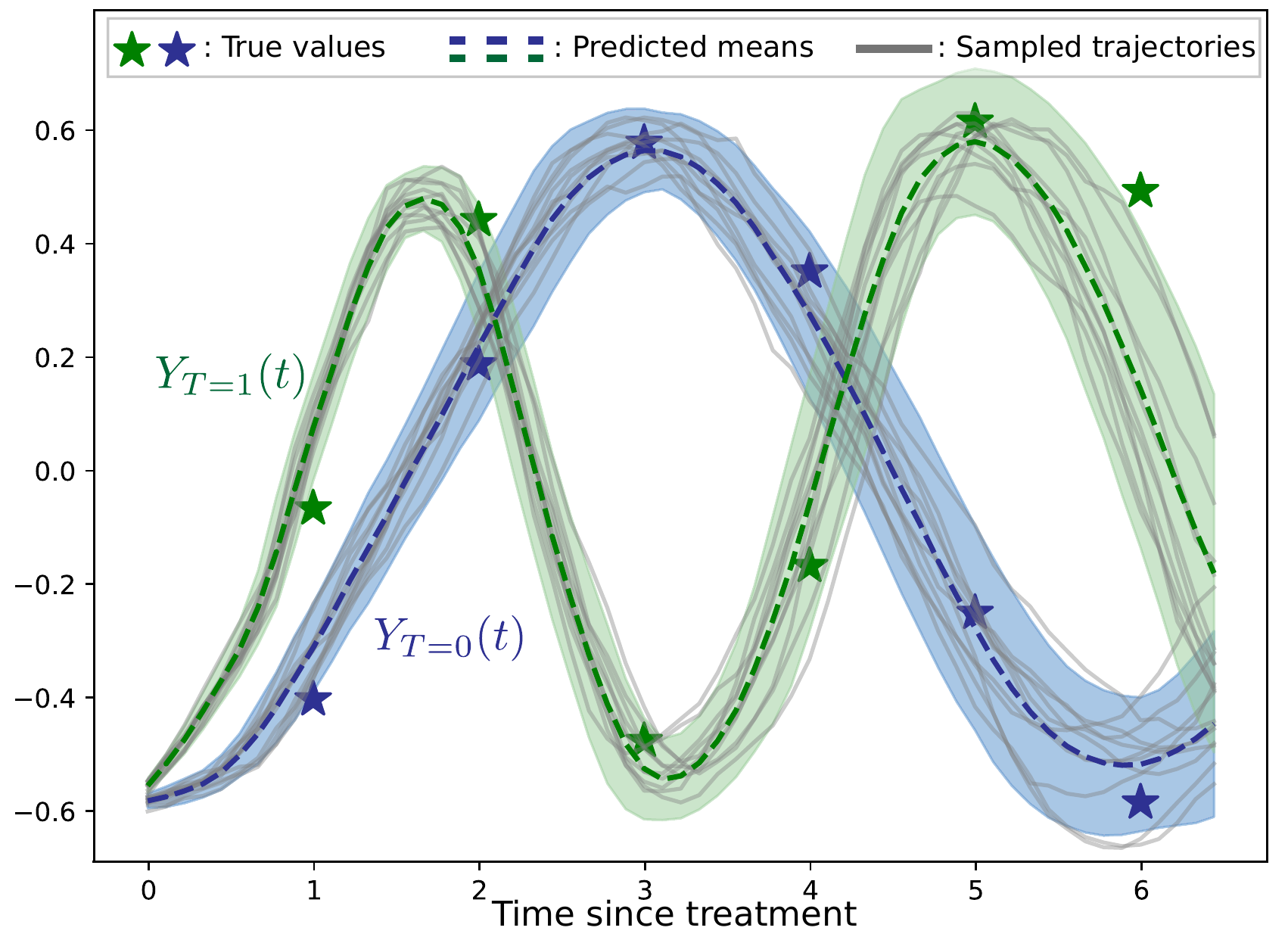}
  \caption{Example of potential outcomes predictions on the oscillator dataset. Means and standard deviations are computed with 50 sampled trajectories, and ten are shown.}
  \label{fig:example_trajec}
\end{figure}

\subsection{Factual And Counterfactual Predictions}

Table \ref{tab:results} summarizes the results from \method against the considered baselines in terms of RMSE on factual, counterfactual and treatment effect estimation. We see that our approach is competitive in all considered metrics and outperforms competitors for both the oscillator and dexamethasone datasets. We observe that the GP performs poorly in almost all datasets, which can be explained by the restrictive additive assumption of the treatment effect (we refer to Appendix \ref{app:baselines} for more details). In Figure \ref{fig:example_trajec}, we show an example of predictions of our model on the oscillator dataset on two distinct potential outcomes.

\begin{figure}[ht]
    \centering
    \includegraphics[width=0.48\textwidth]{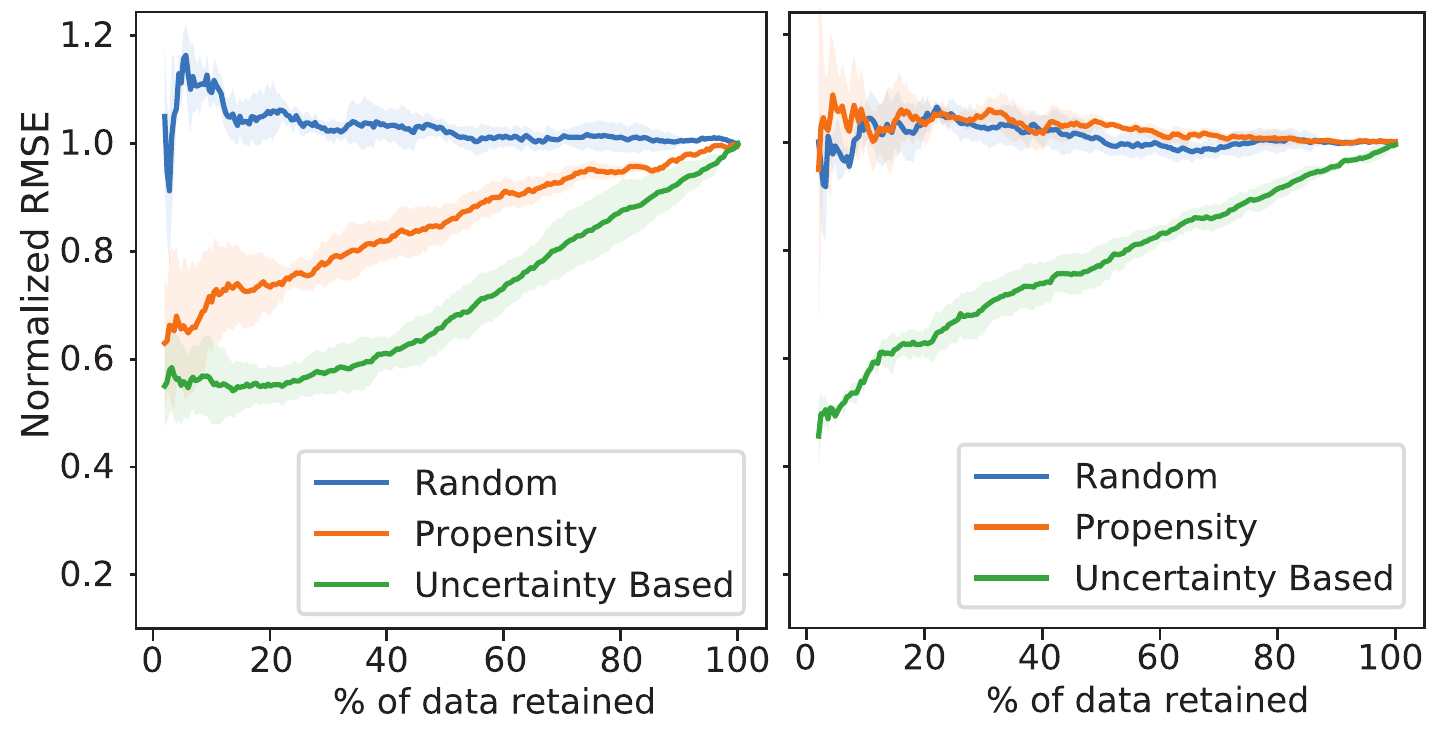}
  \caption{Evolution of normalized PEHE with the number of datapoints retained in the test set for different data pruning strategies on the harmonic oscillator dataset. Left : $\gamma=8$. Right : $\gamma=0$ (no confounding).}
  \label{fig:pehe_curves}
\end{figure}

\subsection{Using ODE Uncertainties To Assess When Counterfactuals Can Be Predicted}
\label{sec:thining}

As stated above, we use the uncertainty of the \method to assess when potential outcomes predictions are reliable. Lower uncertainties should correspond to higher probability of accurate reconstruction and hence lower error (RMSE). We assess the value of the uncertainty estimates as in \citet{jesson2020identifying}. In Figure \ref{fig:pehe_curves}, we show the relative improvement in PEHE that can be obtained with \method if we filter out datapoints according the predicted uncertainty of the model. 
We compare this approach against the random  and propensity-based strategy. 
As observed on Figure \ref{fig:pehe_curves} (Left), the overall PEHE is significantly reduced using the uncertainty of the model, highlighting the usefulness of the uncertainty estimates.

We observe on Figure \ref{fig:pehe_curves} that the uncertainty based approaches and the propensity score based approach are significantly better than the random strategy when confounding is present ($\gamma=8$), suggesting that uncertainties and propensity scores are correlated, as expected. Indeed, extreme propensities correspond to regions in the data with low overlap between treated and untreated cases. Yet, we observe that uncertainties are more effective at selecting samples than the propensity scores. We posit that this is because lack of overlap is not the only source of uncertainty. For instance, lack of similarity, \emph{i.e.} when similar data points are unrepresented in the dataset is another source of uncertainty of the model. Unlike propensity scores, our approach thus appears able to detect other sources of uncertainties, a phenomenon reported in previous studies \citep{jesson2021quantifying}. Indeed, in case of no confounding ($\gamma=0$), propensities do not provide any information, as all treated and untreated distribution perfectly overlap. In contrast, using uncertainties to select reliable predictions is still effective, as it encodes other sources of uncertainties as well.

Lastly, Table \ref{tab:results} also reports the results of the trimming operation of our method for all datasets and compares it against baselines. We observe that the uncertainty-based strategy is almost always the most beneficial, leading to lower RMSE in all metrics.

\subsection{Using Uncertainties To Improve Treatment Assignment Strategies}
\label{sec:treatment_cost}

As introduced in the previous sections, providing uncertainties about potential outcomes can help clinicians develop better strategies for treatment assignment. We illustrate this with a scenario where the outcome of interest is the effect of a treatment at some specific time in the future on some scalar variable (\emph{e.g.} the level of a vital measurement). In our example, we wish the treatment effect to be positive. 

\begin{figure}[hb]
    \centering
    \includegraphics[width=0.4\textwidth]{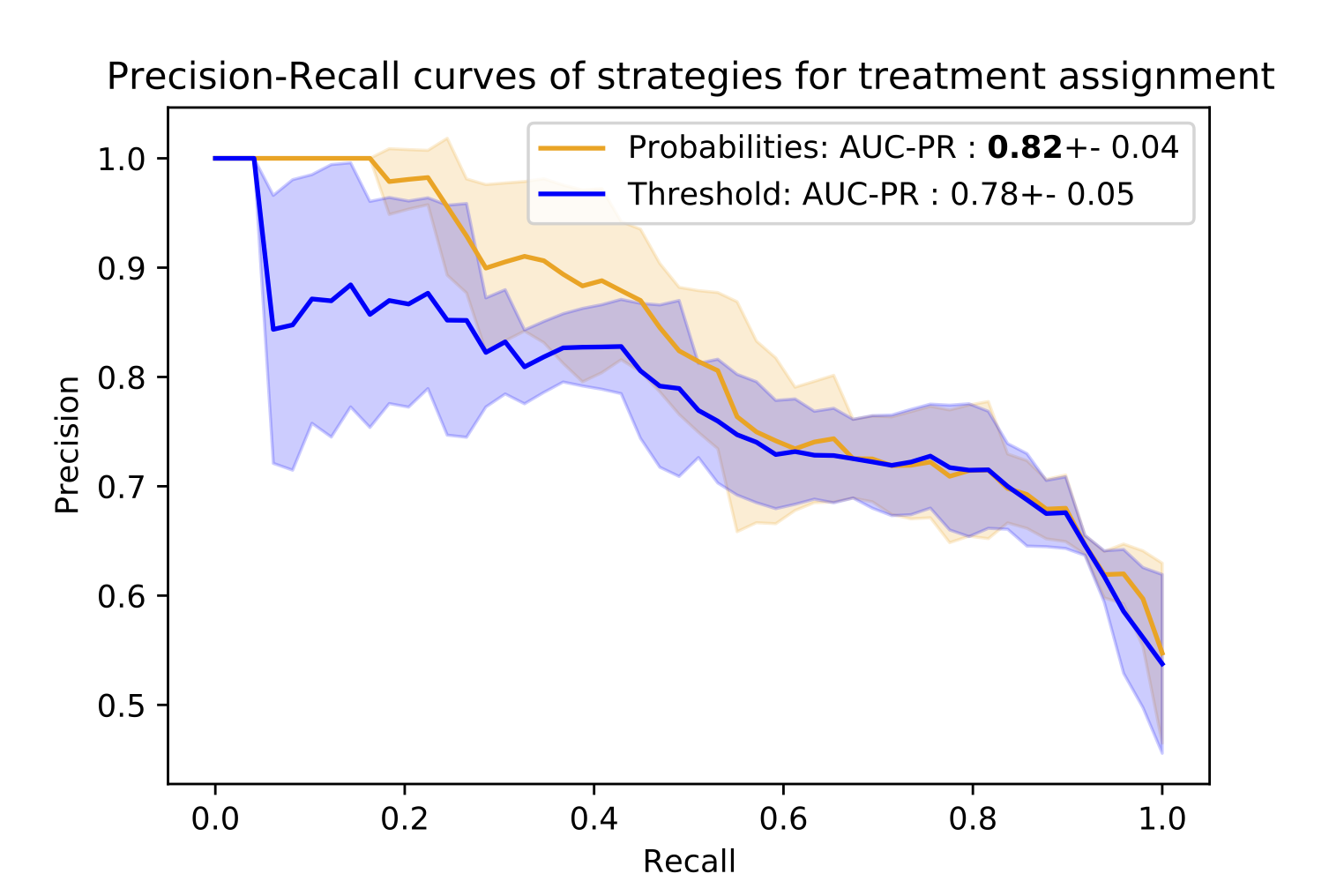}
  \caption{Precision-Recall curves for different different treatment strategies using \method. \emph{Probabilities} refers to a method relying on the uncertainty estimates while \emph{Threshold} only uses the point estimates. 
  }
  \label{fig:pr_strategy}
\end{figure}

Using \method, we can design different strategies for selecting which patients should receive the treatment. First is to use the point prediction of the individual treatment effect and decide to treat if this prediction is higher than a threshold. However, as the model includes uncertainty estimates, one can also use the probability of the individual treatment effect being positive to select the patients to be treated. On Figure \ref{fig:pr_strategy}, we present the precision-recall (PR) curve for both strategies for the oscillator dataset. 
The area under the PR curve (AUC-PR) is significantly higher (pairwise t-test) for the probability-based strategy, with an increase of 4 percentage points. In Appendix \ref{app:treatment_strategies}, we provide another illustration of improved decision making where \method can be used to defer the treatment decision to a clinician when the uncertainty is too large.

In Appendix \ref{app:treatment_strategies}, we also present an experiment where we compare the uncertainty levels between factual and out-of-distribution patients.

\section{RELATED WORK}

\paragraph*{Individualized Treatment Effects.}

Statistical techniques to infer treatment effects from observational data have been an active research area for several decades \citep{pearl2009causality, becker2002estimation}. Among those works, one can distinguish between average treatment effects
 and individual treatment effects (ITE)
 , which is also the focus of this work. 

The majority of the estimators proposed for ITE estimation have mostly focused on the \emph{static} setup (\emph{i.e} without a temporal component), such as propensity scores \citep{lunceford2004stratification}, integral propensity metrics \citep{shalit2017estimating} or counterfactual variance \citep{zhang2020learning}, among others. Those conceptual advances have subsequently been adapted for the \emph{longitudinal} case: marginal structural models \citep{robins2000marginal,lim2018forecasting} extend inverse propensity score weighting and adversarially balanced representations \citep{bica2020estimating} can be seen as an extension of integral propensity metrics. In contrast to our work, those approaches all rely on positive overlap assumption, do not include uncertainty estimates and assume regularly sampled time series. 

\paragraph*{Neural ODEs.}

Our work models the longitudinal treatment outcomes using neural ODEs \citep{chen2018neural}. Since their introduction, they have been successfully used for modelling temporal processes, especially in clinical setting where the sampling is irregular \citep{rubanova2019latent,de2019gru,de2021longitudinal,de2018deep,lee2021learning,yildiz2019ode2vae,ha2018adaptive}. At the intersection with causality, neural ODEs have been exploited for treatment effect modeling \citep{gwak2020neural}, but with no strategy to account for the distribution shift. 
Other works have used neural ODEs for causal \emph{discovery} \citep{bellot2021consistency,de2020latent}, which differs from our goal. 

\paragraph*{Uncertainty Modeling.}

The idea to explicitly model the uncertainties due to confounding in observational data comes from \citet{jesson2020identifying}, where authors used MCDropout \citep{gal2016dropout} to model epistemic uncertainties of a causal variational auto-encoder \citep{louizos2017causal}, operating in the \emph{static} setting. In contrast, our work attempts to model the uncertainties of a neural ODE architecture whose necessary foundations have been laid recently \citep{xu2021infinitely,tzen2019neural,li2020scalable}. Others have proposed Bayesian version of neural ODEs \citep{dandekar2020bayesian,hegde2021bayesian,haussmann2021learning}. However these approaches are often impractical due to their computational complexity (\emph{e.g.} MCMC sampling) or operate in the observation space, in contrast to our method. 


\section{CONCLUSION}


The capacity to anticipate the effect of treatments at the individual level is an considerable challenge, underlying import scientific endeavors such as precision medicine. Our model, \method, an uncertainty-aware neural differential equation model to predict the impacts of treatments over time, opens up a new promising perspective in that direction.

We demonstrated the improved performance of \method on various datasets with respect to the current state of the art methodologies. Importantly, we showed that incorporating uncertainties in the prediction of potential outcomes was crucial for allowing informed treatment decision. In the context of our recurring clinical example, the uncertainties of \method can guide healthcare professionals about when to trust the model to recommend treatments, fostering a synergistic collaboration between clinicians and machine learning models in the clinical practice. 

Modeling the failures of individual treatment effects predictions with uncertainty is an exciting perspective as it relaxes many of the common assumptions made in this context. Yet, several challenges remain before leveraging this type of models into (clinical) practice. In particular, the best ways to endow neural networks with uncertainty is still a active research area. Our model uses the latest advances in this field but significant improvements (especially in out-of-distribution detection) of Bayesian neural networks are still needed. Neural ODEs are also a recent development in the machine learning community and our approach would of course also benefit from advances in this field. For simplicity, we made some restrictive assumptions such as binary treatment assignments and fixed time of treatment. While our approach does not preclude the more general case (as detailed in Appendix \ref{app:no_treatment_label}) - the backbone of \method can be adapted to a broad range of practical scenarios - we leave the details of these specific adaptations, as well as the aforementioned open questions as future work.

\subsubsection*{Acknowledgements}
Edward De Brouwer is funded by a FWO-SB grant from the Fonds Wetenschappelijk Onderzoek. Edward also thanks Maximilian Ilse, Melanie Pradier, Stephanie Milani and Micah Carroll for the scientific discussions and tap dancing lessons.


\bibliography{camera_ready}

\clearpage

\onecolumn

\appendix

\section{BASELINES IMPLEMENTATION DETAILS}
\label{app:baselines}
In this section, we give additional detail regarding the implementation of the baselines used in the experiments.

\subsection{GP}

The GP baseline is inspired by the model proposed by \citet{schulam2017reliable}. Because the code of this paper is not publicly available, we reimplemented as closely as possible to the model described in the original paper. We make our implementation available in our codebase. 

We use a a composition of a radial basis function kernel, a periodic kernel and a white noise kernel. Based on the available time series $X(t)$ and the observed potential outcomes, we train the parameters of the kernels that are shared for all time series. As in \citet{schulam2017reliable}, we model the treatment effect as a learnable additive term in the mean of the gaussian process. In contrast to the original paper, that uses simple exponential functions, we parametrize the impact on the mean with a fully connected neural network with 3 layers of 50 units each. We train the resulting model using GPytorch \citep{gardner2018gpytorch}.

The additive term in the mean being very restrictive, this model therefore does not allow for enough flexibility to fit the complex trajectories we investigate in this paper, resulting in poor performance as shown on Table \ref{tab:results}.

\subsection{CFR}

We reused the code and hyperparameters made available by the authors and translated into a pytorch lightning module \citep{Falcon_PyTorch_Lightning_2019}, to ease reproducibility and comparison between the different models. The resulting implementation is also available in our codebase.

\subsection{IMODE}

We used the code made available by the authors and added some modifications to evaluate counterfactuals. The changes we brought to the original implementations are also available in our codebase.

\section{DEALING WITH TIME SERIES WITHOUT TREATMENT ASSIGNMENT}
\label{app:no_treatment_label}

For sake of simplicity, we assume in our experiments that the time of treatment $t^*$ is constant. For instance, this would fit an example where patients in the ICU are given (or not given) a treatment a specific amount of time after admission.
If this simplified assumption can fit various real-world cases, it might not be always be realistic. In this case, we propose a more general training scheme that acommodates arbitrary treatment times. 

For trajectories for which a treatment is observed, the training remains unchanged. For each of those trajectory, we embed the available information $\mathcal{S}_{t^*}(X)$ to the hidden space ($h(t^*)$) and then predict the future trajectory using a neural differential equation.

For trajectories for which we do not observe a treatment however, we lack a time of treatment $t^*$ and it's therefore unclear when the embedding function $\Phi$ should be applied. For the non-treated instances, we therefore propose to sample a treatment time from the observed empirical distribution of treatment times in the set of trajectories for which a treatment time is available.
Abusing notations, let's denote $\mathcal{T}_{T^* \neq 0}$ the empirical distribution of treatment times, we then minimize

\begin{align*}
    \mathcal{L}(\mathcal{D},\theta,\phi) &= \\ \mathbb{E}_{t^*,q_\theta(\mathcal{H}\mid \mathcal{S}_{t^*})} &[ \log p_{\theta}(Y\mid \mathcal{H})] \\
    -  KL_{q_\theta(\mathcal{H}\mid \mathcal{S}_{t^*})} &(q_\theta(\mathcal{H}\mid \mathcal{S}_{t^*}, \phi) \mid \mid p_0(\mathcal{H}\mid \mathcal{S}_{t^*}, \phi))
\end{align*}

where 
\[
    t^*= 
\begin{cases}
    \sim \mathcal{T} & \text{if } T^* = 0\\
    t^*              & \text{if } T^* \neq 0
\end{cases}
\]

A remaining interrogation is in the dependence of treatment time on the hidden process. Without loss of generality, we consider the general case of a point process with intensity function modulated by the latent process : $\lambda(h(t)) > 0$ such that $P(t^* \in [t_-,t_+]) = 1-e^{-\int_{t_-}^{t_+} \lambda(h(s)) ds}$. In this case, the confounding is addressed the same way as in Lemma \ref{lemma:confounder}, because the same assumptions holds regarding the time of treatment.

\section{ABOUT ASSUMPTION \ref{ass:confounder}}
\label{app:assumption_confounder}

As mentioned in the main text, Assumption \ref{ass:confounder} is less restrictive that it might appear. Let us first briefly recall Takens theorem \citep{takens1981detecting}.

Let $X[t] \in {\mathbb{R} ^{d_X}}$ be generated from a chaotic dynamical system that has a strange attractor $\mathcal{M}$ with box-counting dimension $d_M$, where we define an attractor as the manifold towards which the state of a chaotic dynamical system tends to evolve. The dynamics of this system are specified by a flow on $\mathcal{M}$, $\phi_{(\cdot)}(\cdot) : \mathbb{R} \times \mathcal{M} \rightarrow \mathcal{M}$, where $\phi_{\tau}(\mathcal{M}_t) = \mathcal{M}_{t+\tau}$ and $\mathcal{M}_t$ stands for the point on the manifold at time index $t$. This flow is encoded in the ODE of the system. The observed time series $X[t]$ is then obtained through an observation function $f_{obs}(\cdot)$ :  $X[t] = f_{obs}(\mathcal{M}_t)$. Takens' theorem then states that a delay embedding $\Phi$ with delay $\tau$ and embedding dimension $k$

\begin{align*}
\Phi^{k,\tau}_{\phi,\alpha}(\mathcal{M}_t) &= \\
( &\alpha(\phi_{0}(\mathcal{M}_t)), \alpha(\phi_{-\tau}(\mathcal{M}_t)),\ldots, \alpha(\phi_{-k\tau}(\mathcal{M}_t) ))
\end{align*}

is almost surely an embedding of the strange attractor $\mathcal{M}$ if $k > 2d_M$ and $\alpha: \mathbb{R}^{d_M} \rightarrow \mathbb{R}$ is a twice-differentiable observation function.
More specifically, the embedding map $\Phi$ is a diffeomorphism between the original strange attractor manifold $\mathcal{M}$ and a shadow attractor manifold $\mathcal{M'}$ generated by the delay embeddings. Under these assumptions, one can then theoretically reconstruct the original time series from the delay embedding.

Using this key result, we can conclude that if $t^*$ is large enough (if we have enough history on the time series before prediction), then any twice-differentiable observation function would almost surely provide us with an injective map between the filtration $\mathcal{F}_t^*(X)$ and $h(t^*)$ and Assumption \ref{ass:confounder} would then be satisfied.

The above provides supporting evidence for as why Assumption \ref{ass:confounder} might be often be satisfied in practice. However, two difficulties may still be apparent to the reader. First, the fact that the dynamical system is supposed to be chaotic. We note that all dynamical systems are either periodic, quasi-periodic or chaotic. And that many dynamical systems of interests are actually chaotic \citep{rickles2007simple}. Second, because of the irregular sampling nature of the available time series, an appropriate twice-differentiable observation function $\alpha$ might not be available. This is a topic we are currently investigating, but, in the absence of formal proof, we claim that if enough information is contained in $\mathcal{S}_t^*(X)$, that is, the sampling rate is high enough, then Takens holds as well.

\section{USING UNCERTAINTIES TO IMPROVE TREATMENT ASSIGNMENT STRATEGIES}
\label{app:treatment_strategies}

\begin{figure*}[ht]
\centering
\begin{subfigure}{.33\textwidth}
  \centering
   \includegraphics[width=\linewidth]{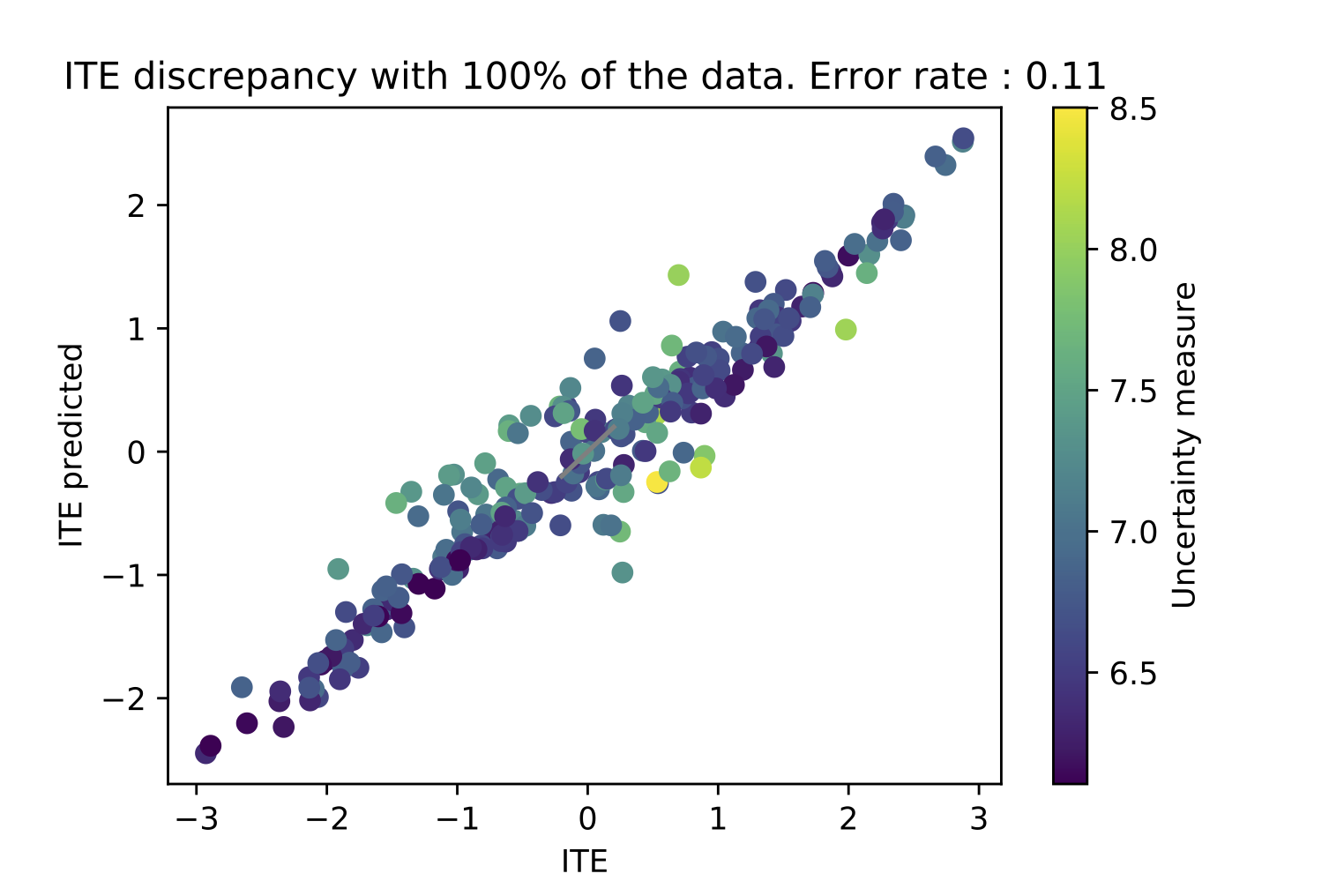}
  \caption{$100\%$ retained}
  \label{fig:ITE_100}
\end{subfigure}%
\begin{subfigure}{.33\textwidth}
  \centering
  \includegraphics[width=\linewidth]{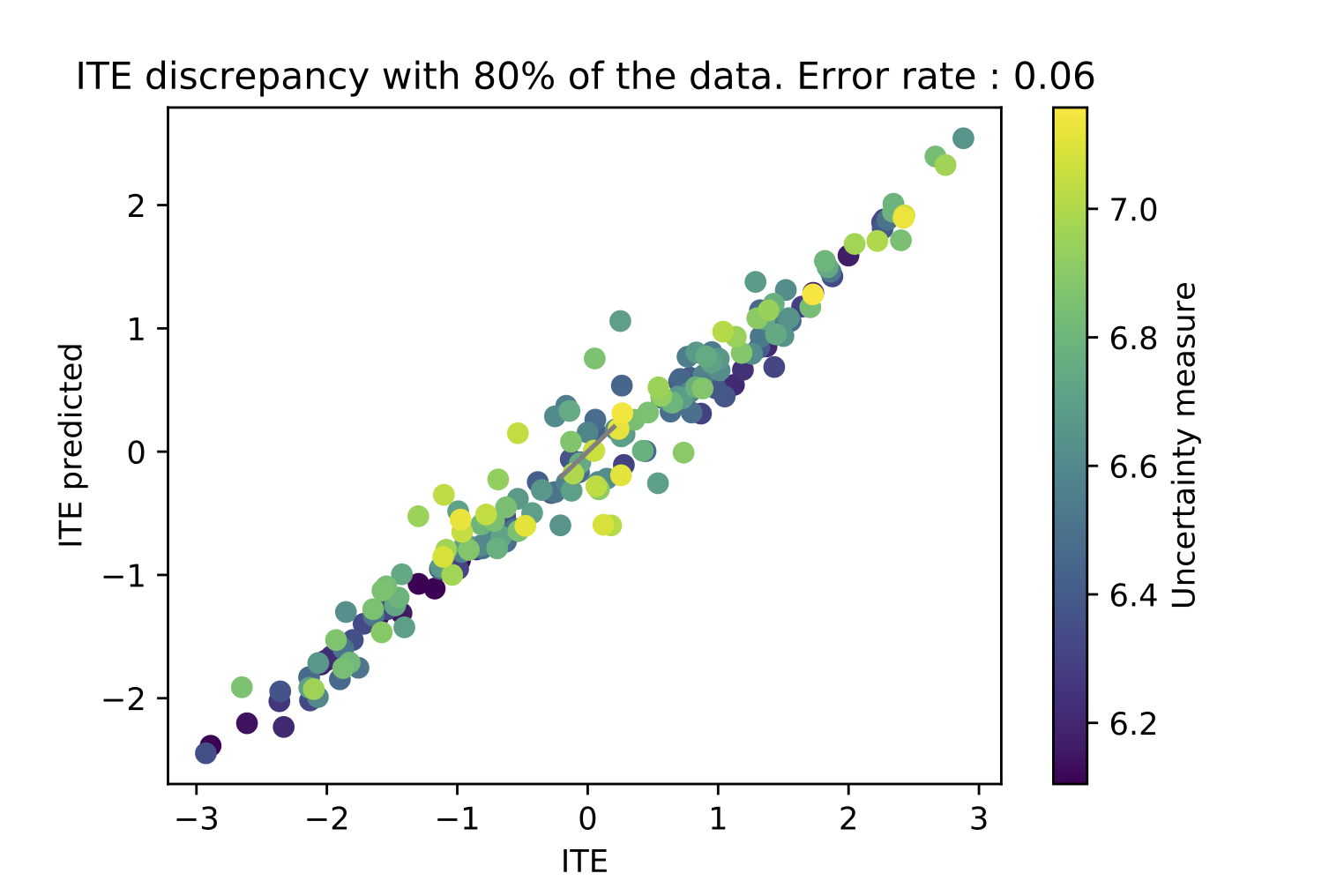}
  \caption{$80\%$ retained}
  \label{fig:ITE_80}
\end{subfigure}
\begin{subfigure}{.33\textwidth}
  \centering
  \includegraphics[width=\linewidth]{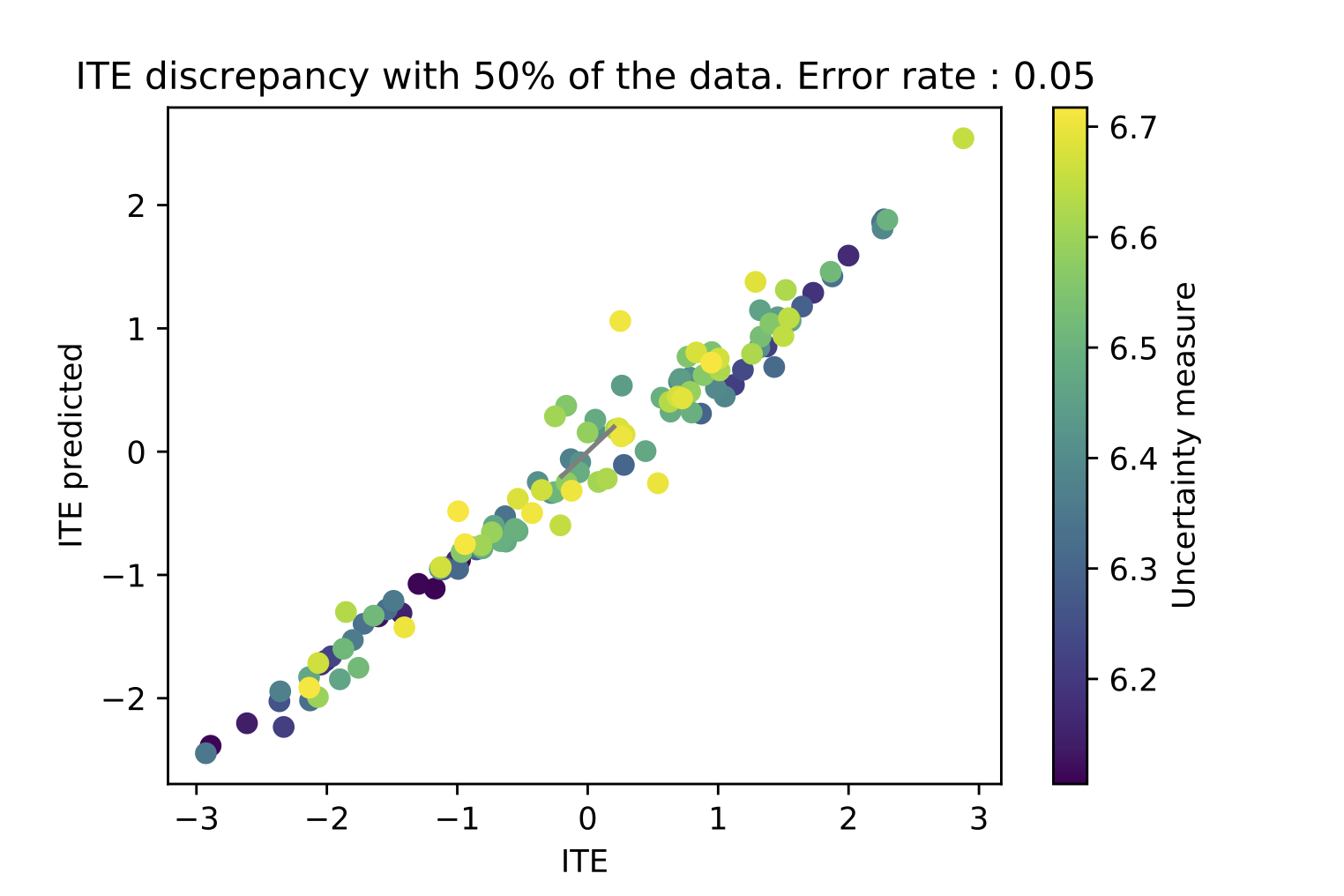}
  \caption{$50\%$ retained}
  \label{fig:ITE_50}
\end{subfigure}
\caption{Evolution of false discovery rate with the percentage of patients used for predictions. Patients with higher uncertainty are discarded first. The x-axis represents the exact individual treatment effect while the y-axis represents the average predictions for each patient. Predictions are getting more accurate (converging on the identity line) and the false discovery rate is decreasing.}
\label{fig:ITE_decreasing}
\end{figure*}

As shown in Section \ref{sec:treatment_cost}, uncertainties can be used to derive probabilities of positive response to treatment that can then inform when to treat a specific patient. In this section, we highlight another approach of using uncertainties to guide treatment decision.

\begin{figure}[ht]
    \centering
    \includegraphics[width=0.48\textwidth]{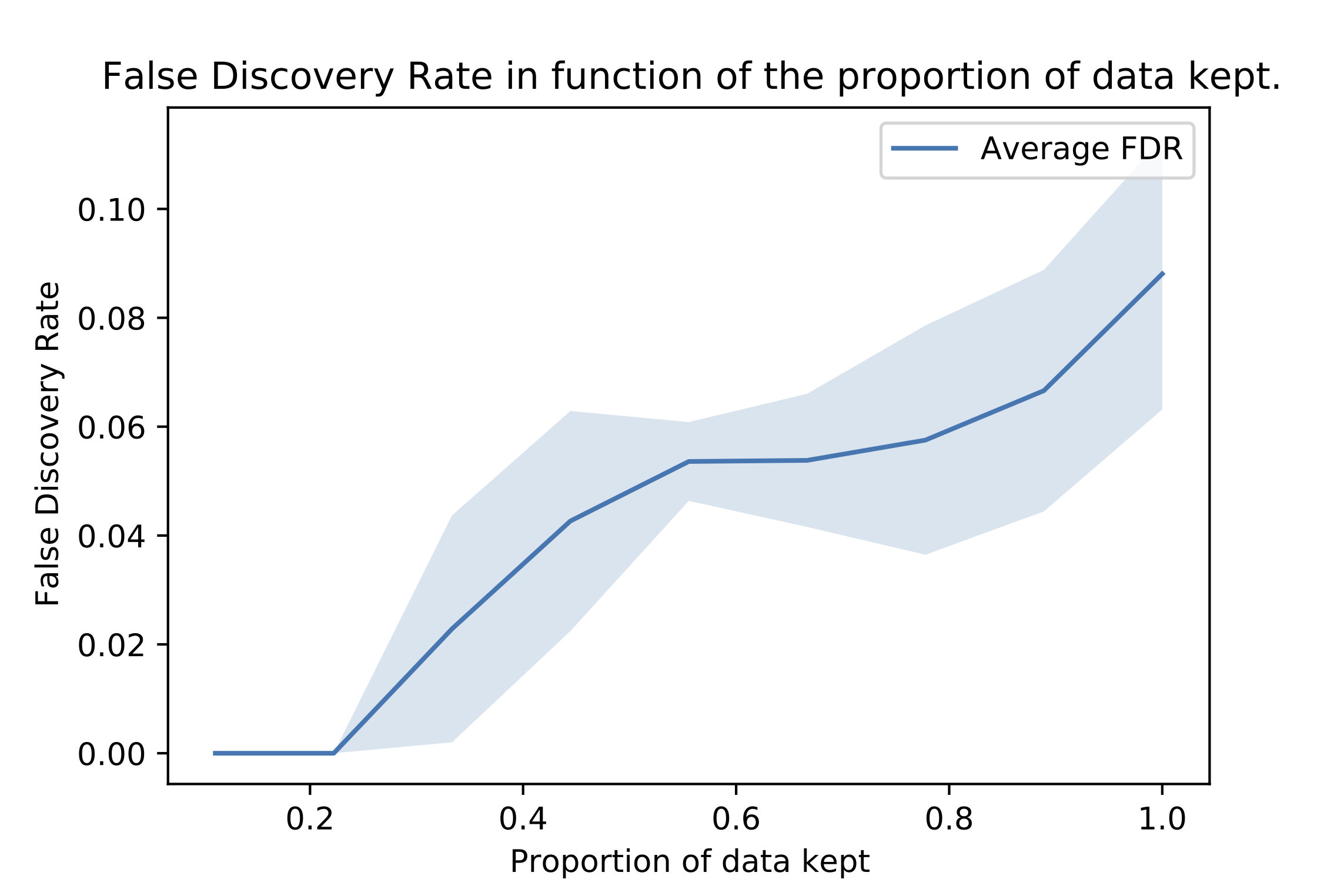}
  \caption{Evolution of the FDR in function of the proportion of data points (patients) kept. Patients with highest uncertainty are discarded first. Shaded are is 2 standard deviations wide.}
  \label{fig:fdr_percent}
\end{figure}

Similarly as in Section \ref{sec:treatment_cost}, we use the oscillator dataset and aim at having a positive treatment effect at some arbitrary time after the giving the treatment. By making a decision of treatment on patients for whom the uncertainty is low, we can decrease the probability of mistakenly assigning a treatment to a patient who will not benefit from it. We frame this error as a false discovery rate ($FDR$) :

\begin{align*}
FDR = \frac{\sum_i^N (ITE_i<0) \cdot (\hat{T}_i)}{\sum_i^N \hat{T}_i}
\end{align*}

where $ITE_i$ is the true individual treatment effect of patient $i$ and $\hat{T}_i \in \{0,1\}$ is the recommendation of treatment that we provide about this patient. We can decide to treat a patient whenever $ITE_i>0$ (\emph{i.e}, when the average predicted ITE is higher than 0). Using uncertainties, we can also decide to treat only patient for whom the uncertainty is not higher than a certain threshold. On Figure \ref{fig:ITE_decreasing}, we display the average predictions of the model for each patient against the true treatment effects in function of the proportion of data used for predictions. As in Section \ref{sec:thining}, we remove data points with higher uncertainty first. We see that if we focus on the patients with the lowest uncertainties, the predictions get more accurate: they converge towards the optimal identity line (\emph{i.e} the prediction is equal to the true value). Importantly, this also results in lower $FDR$, thus in a lower proportion of patients being treated and who would eventually not benefit from the treatment. On Figure \ref{fig:fdr_percent}, we show the evolution of the FDR in function of the proportion of data kept over the different folds. This rate decreases monotonically as patients with lowest uncertainties are kept.

\begin{figure}[ht]
    \centering
\includegraphics[width=0.4\textwidth]{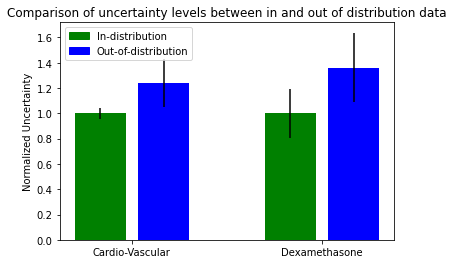}
\caption{Comparison of the normalized uncertainties for in- and out-of-distribution data on the cardiovascular and the dexamethasone dataset.}
\label{fig:ood}
\end{figure}

\subsection{Comparison Between In-distribution and Out-of-distribution data points}

In this experiment, we investigate the uncertainty of our model on out-of-distribution data points. On Figure \ref{fig:ood}, we show the average normalized uncertainty on in-distribution (ID) and out-of-distribution (OOD) data for the cardio-vascular and the dexamethasone data sets. OOD samples here correspond to a type of patients for which no treatment assignment has been observed. As expected, we observe that the uncertainty is significantly larger in the OOD data that in the ID data (pairwise t-test).

\section{ABLATION STUDIES}
\label{app:ablation}

To better understand how the building block of our model contribute individually to the overall performance, we performed an ablation study where we switch off the diffusion part of the model, that would therefore result in a standard Neural ODE. This is referred as the \emph{no diffusion} variant in Table \ref{tab:ablation}.

We observe that the performance degrades significantly, both in the full and the 50\% dataset. This can be explained by the advantageous averaging/ensembling effect of the Bayesian model. What is more, the KL term in the loss function acts as a regularizer of the differential equation, which can contribute to the improvement of performance.

Secondly, we investigate a variant of our model where we $g(h(t))$ outputs both a mean and a standard deviation estimate of the observations $X(t)$ as described in Section \ref{sec:the_method}. We refer to this variant as \emph{with learnable std}. We observe an higher MSE than with the standard version of \method. However, this gap tends to close when the percentage of data points kept decreases. Importantly, despite more appealing theoretical properties, we found that learning the standard deviation of the output distribution made the model harder to train, with instabilities caused by the standard deviation being in the denominator of the log likelihood. 

\begin{table*}[h]
  \caption{%
   Test RMSE for in-distribution, out-distribution and PEHE for the different ablated versions of \method on the harmonic oscillator dataset.
  }
  \label{tab:ablation}
  \vskip 0.1in
  \centering
  \small
  \newcommand{\gr}{\rowfont{\color{gray}}}
  \newcommand{\NA}{---}
  \sisetup{
    detect-all           = true,
    table-format         = 2.1(2),
    separate-uncertainty = true,
    mode                 = math,
    table-text-alignment = center,
    tight-spacing,
  }
  \robustify\bfseries
  \renewrobustcmd{\bfseries}{\fontseries{b}\selectfont}
  \renewrobustcmd{\boldmath}{}
  \let\b\bfseries
  \setlength{\tabcolsep}{3.0pt}
  
  \begin{tabu}{llSSS}
    \toprule
    {} & \textsc{Method}     & {In-distribution RMSE} & {Out-distribution RMSE} & {PEHE}\\
    \midrule
     $\gamma = 8$   & \method   &\b\num{0.04 \pm 0.01} & \b\num{0.06 \pm 0.01} & \b\num{0.07 \pm 0.01} \\
     & \method no diffusion &  \num{0.17 \pm 0.10} & \num{0.28 \pm 0.17} & \num{0.38 \pm 0.25} \\
    & \method with learnable std  &  \num{0.08 \pm 0.04} & \num{0.12 \pm 0.05} & \num{0.16 \pm 0.08} \\
    
     \midrule
    
    & \method 50 \% & \b\num{0.03 \pm 0.01} & \b\num{0.03 \pm 0.01} & \b\num{0.05 \pm 0.01} \\
    & \method 50 \% no diffusion &  \num{0.11 \pm 0.14} & \num{0.24 \pm 0.28} & \num{0.34 \pm 0.39} \\
     & \method with learnable std 50\%        &  \b\num{0.03 \pm 0.01} & \num{0.05 \pm 0.01} & \num{0.07 \pm 0.03} \\

    \bottomrule
    
  \end{tabu}

\end{table*}


\section{DETAILS FOR THE DERIVATION OF THE VARIATIONAL LOWER BOUND}
\label{app:long_proof}

\subsection{Incorporating Uncertainties In ODEs Using Stochastic Differential Equations}

In order to provide uncertainties estimates for the predictions, we embed uncertainties in the parameters of the ODE. Adopting the formalism from Bayesian neural networks, we posit a prior on the weights of the neural ODE function $f_{\theta_f}(\cdot)$: $\theta_f \sim P(\theta_f)$ and aim at estimating the posterior of the distribution of the ODE parameters conditionned on the available data $P(\theta_f \mid \mathcal{D})$ where $\mathcal{D} = \{ \mathcal{S}_{t^*},Y \}$. The weights being probabilistic, the ODE therefore effectively becomes a \emph{random} differential equation whose generating process goes as

\begin{align*}
\theta_f &\sim P(\theta_f) \\
     h_T(t) &= h_T(t^*) + \int_{t^*}^{t}
    f(h_T(s),u_{T,\theta_u}(s-t^*), \theta) \cdot ds
\end{align*}

where we wrote the dependence on the weigths $\theta_f$ explicitly ($f$ is then only the structure of the neural network, without the weigths). Crucially, this means that the prior on the weights $\theta_f$ parametrizes a prior distribution on the process $h_T(t)\mid h_T(t^*)$. For brevity of the notations, we refer to this process as $\mathcal{H}$. Using the natural decomposition of our model, we can derive a variational bound for the marginal probability of the available data $\mathcal{D}$, where we make the dependence on the treatment implicit:

\begin{align*}
    \log(p(\mathcal{D})) &= \\
    log \int_{\mathcal{H}} \int_{\theta_f} &p(Y\mid \mathcal{H}) \cdot p(\mathcal{H} \mid \mathcal{S}_{t^*},\theta_f) \cdot p(\theta_f) \cdot d\theta_f \cdot d\mathcal{H}. \\
\end{align*}

The quantity $p(\mathcal{H} \mid \mathcal{S}_{t^*},\theta_f)$ is a dirac function (as every realization of $\theta_f$ will lead to a single realization of $\mathcal{H}$) and the quantity $ \int_{\theta_f} p(\mathcal{H} \mid \mathcal{S}_{t^*},\theta_f) \cdot p(\theta_f) d\theta_f = p_0(\mathcal{H} \mid \mathcal{S}_{t^*})$ corresponds to a prior of the latent process generated by the prior on $\theta_f$. We can then further write

\begin{align}
    \log(p(\mathcal{D})) &= \log \int_h p(Y\mid \mathcal{H}) \cdot p_0(\mathcal{H} \mid \mathcal{S}_{t^*}) \cdot d\mathcal{H} \nonumber \\
    &= \log \int_h p(Y\mid \mathcal{H}) p_0(\mathcal{H} \mid \mathcal{S}_{t^*})  \frac{q_\theta(\mathcal{H}\mid \mathcal{S}_{t^*})}{q_\theta(\mathcal{H}\mid \mathcal{S}_{t^*})} \cdot d\mathcal{H} \nonumber\\
    & \leq \mathbb{E}_{q_\theta(\mathcal{H}\mid \mathcal{S}_{t^*})} \left[ \frac{\log p(Y\mid \mathcal{H}) \cdot p_0(\mathcal{H}\mid \mathcal{S}_{t^*})}{q_\theta(\mathcal{H}\mid \mathcal{S}_{t^*})}\right] \nonumber \\
    &=  \mathbb{E}_{q_\theta(\mathcal{H}\mid \mathcal{S}_{t^*})} [ \log p(Y\mid \mathcal{H})] \nonumber\\ 
    &\quad - KL_{q_\theta(\mathcal{H}\mid \mathcal{S}_{t^*})}(q_\theta(\mathcal{H}\mid \mathcal{S}_{t^*}) \mid \mid p_0(\mathcal{H}\mid \mathcal{S}_{t^*}))
\end{align}

The process $\mathcal{H}$ being stochastic, we parametrize the posterior distribution $q_\theta(\mathcal{H}\mid X)$ with a stochastic differential equation, in particular a diffusion process:

\begin{align}
    q_\theta(\mathcal{H}\mid \mathcal{S}_{t^*}) \sim d\mathcal{H} = f_{\theta_f}(h(t),u_{T}(t-t^*)) dt + g_\phi(h(t)) dW_t
\end{align}

In order to have a tractable KL divergence term in equation \ref{eq:variational}, we choose the prior of process h(t) to be a diffusion process with same diffusion parameter $g_\theta$. This requirement appears a priori very restrictive but any posterior can be approximated arbitrarily closely by such a functional form given a sufficiently expressive drift process \citep{boue1998variational,tzen2019neural,xu2021infinitely}. The prior on the process $h(t)$ then writes : 

\begin{align*}
    p_0(\mathcal{H}\mid X) \sim dh(t) = f_\phi(h(t)) dt + g_\phi(h(t)) dW_t,
\end{align*}

The KL divergence is then tractable and given by 

\begin{align*}
   KL_{q_\theta(h\mid X)}(q_\theta(h\mid X) &\mid \mid p_0(\mathcal{H}\mid X)) \\
   &= \mathbb{E}_{q_\gamma(\mathcal{H}\mid X)} [\int_0^T \mid \mid u(t,\gamma) \mid \mid_2^2 dt] \\
   \text{where} \quad u(t,\gamma) &=  g_\phi(h(t))^{-1} [f_\gamma(h(t))-f_\phi(h(t))]
\end{align*}

\section{DATASETS}
\label{app:datasets}

\subsection{Harmonic Oscillator}
\label{app:data_pendulum}

We first demonstrate the performance of our approach on a synthetic dynamical system. We model the dynamics of a pendulum where the intervention consists in injecting energy in the system over time by increasing the velocity of the pendulum weight. The structural equations are given in Equation \ref{eq:pendulum}. We further consider that only the angle of the pendulum is observed (the complete state is thus never fully observed) and that observations are made at random and irregularly over time. Different trajectories are generated by sampling different initial angles $\theta_0$ and lengths $l$ : $\theta_0 \sim \mathcal{U}(0.5,1.5)$ and $l \sim \mathcal{U}(0.5,4.5)$, resulting in different trajectories amplitudes and frequencies. Treatment assignments are coupled with the trajectories by setting $P(T=0 \mid \theta_0) = \sigma(\gamma(\theta_0-1))$ where $\sigma(\cdot)$ is the \emph{sigmoid} function and $\gamma$ is a parameter tuning the degree of confounding. The dosage of the treatment $A$ is set as a function of the the amplitude as well.

\begin{align}
\label{eq:pendulum}
    \frac{d\theta(t)}{dt} &= v \nonumber \\ 
    \frac{dv(t)}{dt} &= (1+u(t))\left(\frac{-g}{l}\right)\sin(\theta(t))   \\
    u(t) &=  A\sin(\phi t)\cdot e^{-\delta t} \nonumber
 \end{align}

\subsection{CardioVascular Model : Assessing The Impact Of Fluids Intake}
\label{app:data_cv}

We use a model of the cardiovascular system as proposed in \citep{zenker2007inverse,linial2021generative} and use to study the capacity of our model to learn the impact of fluids intake. Fluid intake is commonly used for treating severe hypotension. However, the response of patient to fluids intake is difficult to assess beforehand. In particular, it depends on the patients cardiac contractility factor and the blood pressure at time of injection. If blood pressure is commonly and easily measured in standard clinical practice, assessing the cardiac contractility level of a patient requires imaging techinques such as echocardiography to measure the stroke volume. Yet, the injection of significant volume of fluids in an irreponsive patients cardiovascular system can lead to severe damage. This lead some clinicians to advocate for fluid challenges, or limited amount of fluid injection to test the responsiveness. This technique is still contested in the medical community and legs raising challenge, much less damaging but also less effective at assessing a patients response has been encouraged. This lack of availability of clear guidelines for fluids intake makes it a perfect case study for counterfactual prediction. Indeed, we'll try to address the question of if a clinician should administer fluids to particular patient based on his clinical history and therefore help informing clinical practice. The system of ODE used to generate the data is the following :

\begin{align*}
\frac{d S V(t)}{d t} &=I_{\text {external }(t)} \\
\frac{d P_{a}(t)}{d t} &=\frac{1}{C_{a}}\left(\frac{P_{a}(t)-P_{v}(t)}{R_{T P R}(S)}-S V \cdot f_{H R}(S)\right) \\
\frac{d P_{v}(t)}{d t} &=\frac{1}{C_{v}}\left(-C_{a} \frac{d P_{a}(t)}{d t}+I_{\text {external }(t)}\right) \\
\frac{d S(t)}{d t} &=\frac{1}{\tau_{\text {Baro }}}\left(1-\frac{1}{1+e^{-k_{\text {width }}\left(P_{a}(t)-P_{a_{\text {set }}}\right)}}-S\right)
\end{align*}

where

\begin{align*}
R_{T P R}(S) &=S(t)\left(R_{T P R_{M a x}}-R_{T P R_{M i n}}\right)
\\ &\quad +R_{T P R_{M i n}}+R_{T P R_{M o d}} \\
f_{H R}(S) &=S(t)\left(f_{H R_{M a x}}-f_{H R_{M i n}}\right)+f_{H R_{M i n}} .
\end{align*}

In the above dynamical system, $P_a, P_v, S and SV$ stand for arterial blood pressure, venous blood pressure, autonomic baroreflex tone and cardiac stroke volume respectively. $I_{\text {external }(t)}$ is the amount of fluids given the patient over time and corresponds to the exogeneous input $u_T(t)$ in our model. In the data generation, we model it as 

\begin{align*}
    I_{\text {external }(t)} = 5*e^{-(\frac{t-5}{5})^2}
\end{align*}

The treatment assignment is confounded with the history of the patient and we make it explicity depends on the value of the arterial blood pressure at the time of treatment : 

\begin{align*}
    P(T=1) = \sigma(\gamma \cdot (\frac{P_a(t^*)-P_{a,min}}{P_{a,width}}-0.5))
\end{align*}

where $\sigma$ is the sigmoid function and $P_{a,min}$ and $P_{a,width}$ are defined parameters (75 and 10)).

\subsubsection{Pharmacodynamics Model}
\label{app:data_pharma}

On top of the previous datasets, we also consider the pharmacodynamics of a dexamethasone, a glucocorticoid drug that has been used in treatment against COVID19. The dynamical system, presented in Equation \ref{eq:covid} is adapted from \citet{dai2021prototype} and \citet{qian2021integrating}. Variables $z_1$ and $z_5$ represebt the innate and adapative immune response, $z_2$ and $z_3$ the concentration of dexamethasone in the lung tissue and plasma and $z_4$ represents the viral load. 

\begin{align}
\label{eq:covid}
\dot{z}_{1} &=k_{I R} \cdot z_{4}+k_{P F} \cdot z_{4} \cdot z_{1}-k_{O} \cdot z_{1} \nonumber \\
 &\quad +\frac{E_{\max } \cdot z_{1}^{h_{P}}}{E C_{50}^{h_{P}}+z_{1}^{h_{P}}}-k_{D e x} \cdot z_{1} \cdot z_{2} \\
\dot{z}_{2} &= -k_2 \cdot z_2 + k_3 \cdot z_3 \\
\dot{z}_{3} &= -k_3 \cdot z_3 \\
\dot{z}_{4} &=k_{D P} \cdot z_{4}-k_{I I R} \cdot z_{4} \cdot z_{1}-k_{D C} \cdot z_{4} \cdot z_{5}^{h_{C}} \\
\dot{z}_{5} &=k_{1} \cdot z_{1}
\end{align}

Only variables $z_1$ and $z_5$ can be realistically measured in the lab, through Type I IFNs and Cytotoxic T Cells respectively. Therefore we only use those two variables in $X(t)$. We set the variable $z_1$ as the variable of interest for the treatment effect. We model the intervention by simulating a constant injection of dexamethasone in the plasma ($\dot{z_3} = 10$). We introduce confounding by modeling a dependence of the treatment assignment on the factor $k_{Dex}$, that modulates the impact of dexamethasone on the immune response. We set $ P(T=1) =\sigma(\frac{k_{Dex-1}}{15} -0.5)$ with $K_{Dex} \sim \mathcal{U}(1,16)$.

\end{document}